\documentclass[letterpaper, 10 pt, conference]{ieeeconf}  
\usepackage{graphicx} 
\usepackage[hyphens]{url}  
\usepackage{algorithm}
\usepackage{algorithmic}

\usepackage{amsmath,amssymb,amsfonts}
\usepackage{makecell}
\usepackage{subfigure} 
\usepackage{stfloats}
\usepackage{newfloat}
\usepackage{listings}
\usepackage{hyperref}
\usepackage[T1]{fontenc}
\usepackage{aecompl}
\usepackage{booktabs}
\usepackage{bbding}
\usepackage{multirow} 
\usepackage{svg} 
\usepackage{threeparttable}
\usepackage{url}
\usepackage{cleveref}
\usepackage{cite}
\usepackage{footmisc}
\IEEEoverridecommandlockouts                              

\title{\LARGE \bf
Bidirectional Task--Motion Planning Based on Hierarchical Reinforcement Learning for Strategic Confrontation
}

\author{Qizhen Wu$^{1}$, Lei Chen$^{2}$, Kexin Liu$^{3}$, and Jinhu L\"u$^{4}$
\thanks{*This work was supported in part by the National Key Research and Development Program of China under Grant 2022YFB3305600, and in part by the National Natural Science Foundation of China under Grants 62141604, 62088101, and 62003015. (Corresponding author: Lei Chen)}
\thanks{$^{1}$Qizhen Wu, $^{3}$Kexin Liu, and $^{4}$Jinhu L\"u are with the School of Automation Science and Electrical Engineering,
        Beihang University, Beijing 100191, China.
        ({wuqzh7@buaa.edu.cn, skxliu@163.com, jhlu@iss.ac.cn})}%
\thanks{$^{2}$Lei Chen is with the Advanced Research Institute of Multidisciplinary Sciences and State Key Laboratory of CNS/ATM, Beijing Institute of Technology, Beijing 100081, China.
        ({bit$\_$chen@bit.edu.cn})}%
\thanks{The experiment video is avaliable at \url{https://www.bilibili.com/video/BV1JTwmeaEeN/?vd_source=9de61aecdd9fb684e546d032ef7fe7bf}.}%
}

\begin{document}

\maketitle
\thispagestyle{empty}
\pagestyle{empty}

\begin{abstract}
In swarm robotics, confrontation scenarios, including strategic confrontations, require efficient decision--making that integrates discrete commands and continuous actions. Traditional task and motion planning methods separate decision--making into two layers, but their unidirectional structure fails to capture the interdependence between these layers, limiting adaptability in dynamic environments. Here, we propose a novel bidirectional approach based on hierarchical reinforcement learning, enabling dynamic interaction between the layers. This method effectively maps commands to task allocation and actions to path planning, while leveraging cross--training techniques to enhance learning across the hierarchical framework. Furthermore, we introduce a trajectory prediction model that bridges abstract task representations with actionable planning goals. In our experiments, it achieves over 80\% in confrontation win rate and under 0.01 seconds in decision time, outperforming existing approaches. Demonstrations through large--scale tests and real--world robot experiments further emphasize the generalization capabilities and practical applicability of our method.

\end{abstract}

\section{INTRODUCTION}

Recent advances in artificial intelligence lead to significant progress in robotics \cite{guo2024powerful,kannan2024smart}, with particular attention given to robotic swarm confrontations \cite{fan2024deep,piao2023spatio}. In these scenarios, AI--driven decision--making empowers robots to engage in strategic tasks such as pursuit--evasion \cite{wu2024hrl}, defense--attack \cite{liu2024game}, and strategic confrontations \cite{vinyals2019grandmaster}. These applications highlight the need for effective decision--making mechanisms capable of handling a hybrid process, including discrete commands and continuous actions \cite{hou2023hierarchical}, which are central to the success of modern robotic systems. 

Traditional methods for solving such problems rely heavily on task and motion planning (TAMP) \cite{vaquero2024eels}, which divides the problem into two distinct decision--making layers. Although conventional TAMP methods \cite{hou2023hierarchical,wang2023decision} demonstrate effectiveness by organizing tasks within detailed planning frameworks, they suffer limitations due to their reliance on expert knowledge . The complexity of modeling and task specification renders these methods less accessible for non--expert users. Heuristic algorithms \cite{liu2022evolutionary,braun2022rhh} attempt to address these challenges by adaptively searching for feasible solutions, but the independent nature of the hierarchical architecture prevents the upper layer from verifying the enforceability of commands transferred to the lower layer. Furthermore, TAMP methods struggle with generalization, as evidenced by the frequent need for repeated adjustments \cite{xue2024d,ren2024extended} when adapting to dynamic environments or scaling to larger robot populations.

\begin{figure}[t]
    \centering    \includegraphics[width=0.4\textwidth]{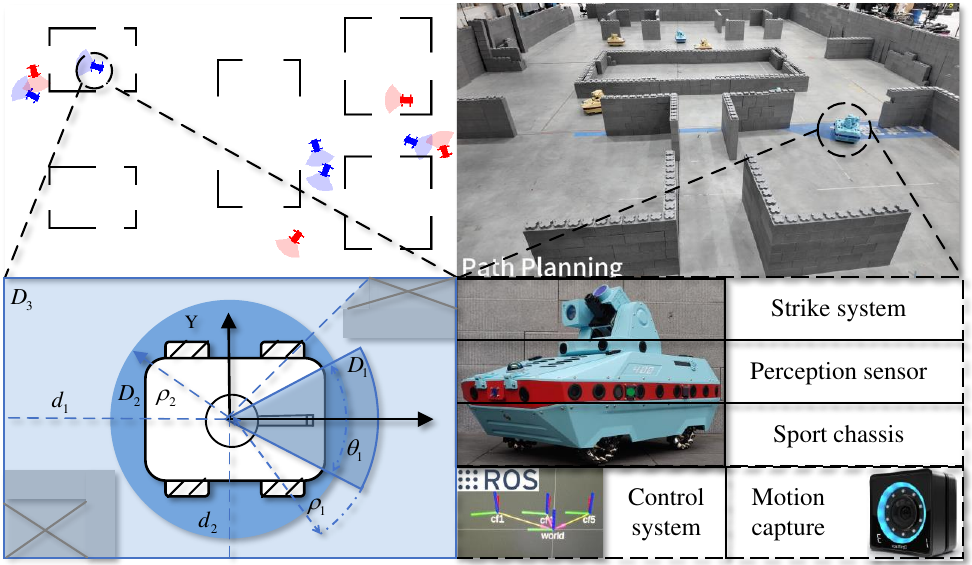}
    \caption{Simulations and real--robot systems in a confrontation scenario.}
    \label{fig1}
\end{figure}

In contrast, deep reinforcement learning (DRL) \cite{kaufmann2023champion,haarnoja2024learning} emerges as a promising alternative by using a single--layer framework \cite{qu2023pursuit} that learns optimal decision--making strategies through end--to--end training. It aims to maximize cumulative rewards, fine--tuning its decision network to produce effective strategies. However, in complex environments \cite{eppe2022intelligent}, DRL encounters challenges due to its focus on global objectives, leading to sparse--reward scenarios \cite{dawood2023handling}. This results in practical limitations when applied to intricate real--world tasks.

To address these limitations, hierarchical reinforcement learning (HRL) \cite{huang2024hilma} provides a more flexible solution by decomposing the complex task into two independently trainable networks, similar to the TAMP approach. It fosters a dynamic feedback loop between the two layers, enabling them to communicate and refine their strategies as they learn. In HRL, the upper layer divides the task into temporal segments and assigns tailored rewards to guide the lower layer. The method allows the upper layer to refine its task decomposition based on the learning progress of the lower layer, offering a more adaptable framework to tackle complex problems \cite{kong2023hierarchical2,nian2024large}. This hierarchical structure enables more detailed problem--solving, particularly in scenarios where discrete commands and continuous actions are closely intertwined.

In this paper, we propose an innovative HRL method to tackle the unique challenges of swarm confrontation. Unlike previous approaches, our method introduces a two--layer interaction mechanism that enhances both the accuracy and stability of decision--making. Specifically, we integrate task allocation (for discrete commands) and path planning (for continuous actions) within a bidirectional closed--loop architecture. This structure enables effective communication between the two layers, ensuring the enforceability of assigned tasks. Additionally, we introduce a trajectory prediction model that bridges the gap between abstract task representations and executable planning targets, such as predicting the most likely location of enemies in an evasion task. Our experimental results demonstrate that our approach outperforms baseline methods, including traditional TAMP approaches and single--layer DRL, especially in terms of generalization across different scales. Through ablation studies, we highlight the critical importance of the interaction mechanism and show that our method exhibits superior adaptability in real--robot systems illustrated in Fig. \ref{fig1}.

We organize the rest of the paper as follows. Section \ref{sec:sample2} outlines the preliminaries of goal--conditioned HRL and formulates the strategic confrontation problem. Section \ref{sec:sample3} elaborates on the proposed HRL method and its implementation. Section \ref{sec:sample4} presents the experimental evaluations of our approach. Finally, Section \ref{sec:sample5} concludes the paper.
\section{Preliminaries and Problem Description}
\label{sec:sample2}

\subsection{Definition of strategic confrontation}
\label{sec:sample2A}
This study models strategic confrontation as a dynamic process that involves task allocation and path planning in a static--obstacle environment. In this scenario, the blue and red agents compete against each other, with the one that eliminates all opponents being declared the winner. To simplify the agents' movement, we consider the confrontation in two--dimensional space. Let $\boldsymbol{p}$ and $\boldsymbol{v}$ denote the position and velocity vectors in two dimensions, respectively. Agents on both sides are subject to the same constraints on their abilities. For the $i$--th agent $\boldsymbol{u}_i=\left( \boldsymbol{p}_i, \boldsymbol{v}_i, f_i \right)$, we define $\mathbb{U}_a=\left\{ u_{a,j}|j=1,...,N_a \right\}$ and $\mathbb{U}_e=\left\{ u_{e,k}|k=1,...,N_e \right\}$ as the sets of $N_a$ allies and $N_e$ enemies, respectively. Here, $f_i$ represents the task currently performed by the $i$--th agent, defined as searching ($f_i = -3$), escaping ($f_i = -2$), supporting ($f_i = -1$), or chasing ($f_i = 0$). At each time step, every agent assigns a task and then performs path planning based on the task goal to ensure obstacle avoidance.

As shown in Fig. \ref{fig1}, each agent has an attack range $D_1$ modeled as a sector with radius $\rho_1$ and angle $\theta_1$. When an opponent enters the attack range $D_1$, the agent automatically launches an attack. At this point, the probability of the bullet hitting and causing the opponent to die is $\epsilon$. By defining an avoidance range $D_2$ with radius $\rho_2$, the agent plans feasible paths to avoid collisions with obstacles. In this work, the $i$--th agent gathers information about neighboring allies and enemies through communication and perception, respectively. We define the communication and perception ranges as a rectangle with length $d_1$ and width $d_2$. While obstacles obstruct perception, they do not interfere with communication. When the $j$--th agent is within the communication and perception ranges of the $i$--th agent, we label it as $j \in \mathbb{N}_i$, where $\mathbb{N}_i$ denotes the set of agents observable by the $i$--th agent. This confrontation setting provides the foundational context for the goal--conditioned HRL framework.

\begin{figure*}[t]
    \centering
    \includegraphics[width=0.9\textwidth]{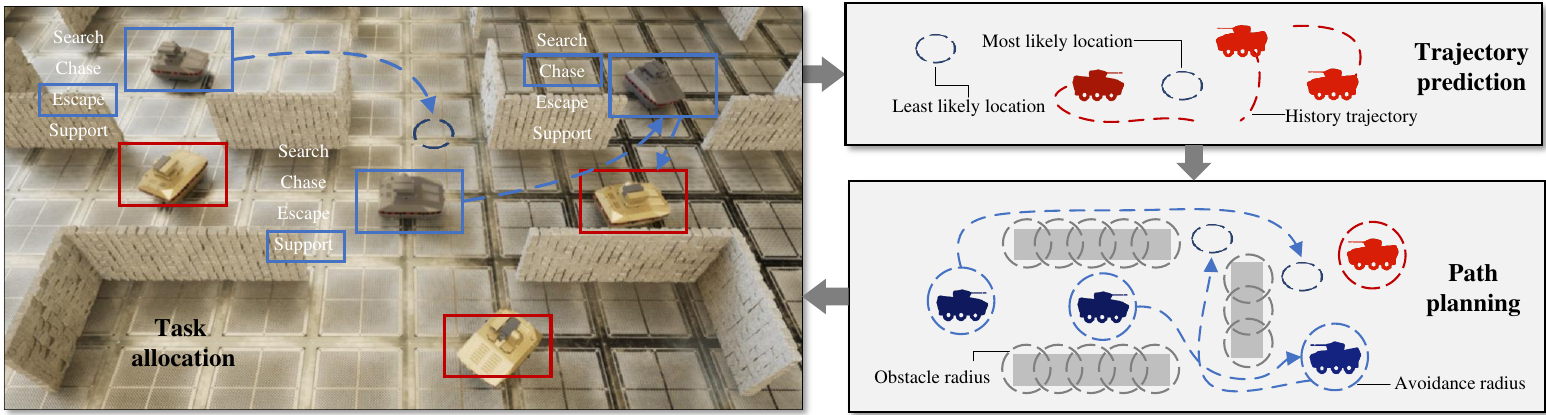}
    \caption{Overview of the proposed framework integrating task allocation (left), trajectory prediction (top right), and path planning (bottom right) modules.}
   \label{fig2}
\end{figure*}

\subsection{Goal--conditioned HRL}
\label{sec:sample2B}
Let $\mathbb{R}$ denote the set of real numbers, $\mathbb{R}^{+}$ the set of positive real numbers, and $\mathbb{Z}$ the set of integers. We formulate a long--horizon planning task for $N$ agents as a goal--conditioned Markov Decision Process (goal--conditioned MDP), represented by the eight--tuple $\left( N, \mathbb{S}, \mathbb{G}, \mathbb{O}, \mathbb{A}, P, R, \gamma \right)$, where:
\begin{itemize}
\item $\mathbb{S} \subseteq \mathbb{R}^{m_s}$, $\mathbb{A} \subseteq \mathbb{R}^{m_a}$, $\mathbb{O} \subseteq \mathbb{R}^{m_o}$, and $\mathbb{G} \subseteq \mathbb{R}^{m_g}$ represent the state space, the action space, the observation space, and the goal space, respectively.
\item $m_s$, $m_a$, $m_o$, and $m_g$ denote the dimensions of the state, action, observation, and goal spaces, respectively.
\item $\boldsymbol{s}\in \mathbb{S}$ represents the state of the environment.
\item $\boldsymbol{z}_i\in \mathbb{O}$ represents the observation of the $i$--th agent and $\boldsymbol{z} = \left( \boldsymbol{z}_1,\cdots , \boldsymbol{z}_N \right) $ denotes the joint observations.
\item $\boldsymbol{a}_i\in \mathbb{A}$ represents the action of the $i$--th agent and $\boldsymbol{a} = \left( \boldsymbol{a}_1,\cdots ,\boldsymbol{a}_N \right)$ denotes the joint actions.
\item $P\left( \boldsymbol{s},\boldsymbol{a}  \right) :\mathbb{S}\times \mathbb{A}\times \mathbb{S}\rightarrow [0,1]$ denotes the state transition probabilities from $\boldsymbol{s}\left(t\right)$ to $\boldsymbol{s}\left(t+1\right)$ under $\boldsymbol{a}\left(t\right)$.
\item $R\left( \boldsymbol{s},\boldsymbol{a}, \boldsymbol{g} \right) :\mathbb{S}\times \mathbb{A}\times \mathbb{G}\rightarrow \mathbb{R}$ is the reward function.
\item  $\gamma \in \left( 0,1 \right)$ represents the discount factor.
\end{itemize}
The subgoal $\boldsymbol{q}$ and the goal $\boldsymbol{g}$ are defined in the goal space $\mathbb{G}$, which is considered a subspace of $\mathbb{S}$, with a mapping function $\mathcal{\varphi } : \mathbb{S} \rightarrow \mathbb{G}$. HRL typically involves two--layer policies: an upper--layer policy $\pi_{u} \left( \boldsymbol{q}_i\left(t\right) \left| \boldsymbol{z}_i\left(t\right), \boldsymbol{g} \right. \right)$ and a lower--layer policy $\pi_{l} \left( \boldsymbol{a}_i\left(t\right) \left| \boldsymbol{z}_i\left(t\right), \boldsymbol{q}_i\left(t\right) \right. \right)$. We assume that each subtask consists of $h$ steps.  The upper layer predicts the subgoal $\boldsymbol{q}_{i,m},m=0,1,\cdots,T_h$, where $T_h=T/h$ and $T$ is the total time. Upon receiving the observation $\boldsymbol{z}_{i,m+n}$, the lower layer attempts to reach the subgoal by executing the appropriate action $\boldsymbol{a}_{i,n}$.  
Starting from the state $\boldsymbol{s}_{m}$, the state--action trajectory for the $m$--th subtask can be represented as $\tau_m= \{ \left( \boldsymbol{s}_{m+n}, \boldsymbol{a}_n, \boldsymbol{s}_{m+n+1} \right) \left| \boldsymbol{s}_{m}, \boldsymbol{q}_{m} \right. \}^{h-1}_{n=0}$.

To ensure the enforceability of the predicted subgoal, the upper--layer policy is guided by the reward feedback from the environment. The reward function and optimization objective in the upper layer can be described as follows:
\begin{equation}
    \begin{array}{c}
        \displaystyle R_u\left( \tau_m \right) =\sum_{n=0}^{h-1}{r_{\text{env}}\left( \boldsymbol{s}_{m+n},\boldsymbol{a}_n,\boldsymbol{g} \right)},
    \end{array}\label{equ1}
\end{equation}
\begin{equation}
    \begin{array}{c}
        \displaystyle
\pi _{u}^{*}=\underset{\pi _{u}}{\arg\max}\sum_{m=0}^{T_h - 1}{\gamma ^m R_u\left( \tau_m \right)}.
\end{array}\label{equ2}
\end{equation}
Here, $\arg\max \nolimits_xF\left( x\right)$ represents the value of the variable $x$ when $F\left( x\right)$ reaches its maximum value. The intrinsic reward function of the lower layer is designed to achieve the subgoal provided by the upper layer. The reward function and optimization objective in the lower layer can be described as
\begin{equation}
    \begin{array}{c}
        \displaystyle R_l\left( \boldsymbol{s}_{m+n},\boldsymbol{a}_n,\boldsymbol{q}_{m} \right) = - D \left( \boldsymbol{s}_{m+n}, \boldsymbol{q}_{m} \right),
    \end{array}\label{equ3}
\end{equation}
\begin{equation}
    \begin{array}{c}
        \displaystyle
\pi _{l}^{*}=\underset{\pi _{l}}{\arg\max}\sum_{n=0}^{T-1}{\gamma ^t R_l\left( \boldsymbol{s}_n,\boldsymbol{a}_n,\boldsymbol{q}_{\left\lfloor n / h \right\rfloor }\right)},
\end{array}\label{equ4}
\end{equation}
where $D : \mathbb{S} \times \mathbb{G} \rightarrow \mathbb{R}^{+}$ is a continuous distance used to characterize the difference between the state $\boldsymbol{s}$ and subgoal $\boldsymbol{q}$. Additionally, $\lfloor x \rfloor =\max \left\{ m\in \mathbb{Z}|m\leqslant x \right\}$, where $x \in \mathbb{R}$.

\section{METHODOLOGY}
\label{sec:sample3}
In this section, we present the HRL method to address the strategic confrontation problem in a static--obstacle environment. An overview of our method is shown in Fig. \ref{fig2}. The lower layer utilizes a multi--agent deep deterministic policy gradient (MADDPG) algorithm \cite{wu2024hrl} for path planning based on subgoals. The upper layer employs a decentralized deep Q--learning (DQN) algorithm \cite{vinyals2019grandmaster} for task allocation. This method integrates the cumulative rewards from the lower layer into the upper layer, establishing an interaction mechanism between the two layers. In addition, we employ a trajectory prediction model to bridge abstract task representations with concrete planning goals. Finally, we introduce a cross--training strategy to enhance the training efficiency and stability of the proposed method.
\subsection{Lower Layer for Path Planning}
\label{sec:sample3A}
We enclose all rectangular obstacles with $N_o$ circles of radius $\rho_3$, where the position of the center of each circle is denoted as $\boldsymbol{p}_{o,c}$, with $c=1,\dots,N_o$. Moreover, we consider allies within the communication range and enemies within the perception range as neighboring agents $\boldsymbol{u}_j$, where $j \in \mathbb{N}_i$. Based on this, we create the following observation:
\begin{equation}
\begin{array}{c}
\boldsymbol{z}_{l,i}=\left(\boldsymbol{p}_i,\boldsymbol{p}_{j},\boldsymbol{v}_i,\{\boldsymbol{v}_{j}\}_{j\in \mathbb{N}_i},\{\boldsymbol{p}_{o,c}\}_{c \in \mathbb{N}_i}\right) .
\end{array}\label{equ5}
\end{equation}
The action consists of the velocity magnitude $\lVert \boldsymbol{v} \rVert \in [0,\lVert \boldsymbol{v} \rVert _{\max}]$ and velocity direction $\psi \in [-\pi, \pi]$ of the $i$--th agent, denoted as
\begin{align}
        a=\left( \lVert \boldsymbol{v}_i \rVert, \psi_i \right),	
    \label{equ6}
\end{align}
where $\lVert \cdot \rVert$ denotes the Euclidean norm and $\lVert \boldsymbol{v} \rVert _{\max}$ denotes the maximum velocity magnitude.
Based on the current state and actions, we can calculate the position at the next time step using the first--order dynamics model
\begin{align}
        p_{x}\left(t+1\right) = p_{x}\left(t\right) + \delta t \lVert \boldsymbol{v} \rVert  \cos \psi, \nonumber \\
        p_{y}\left(t+1\right) = p_{y}\left(t\right)  + \delta t \lVert \boldsymbol{v} \rVert  \sin \psi,
    \label{equ7}
\end{align}
where $p_x$ and $p_y$ denote the horizontal and vertical coordinates of the agent’s position, respectively. The symbol $\delta t$ is the length of each time step.

We design an avoidance reward to guide agents in avoiding collisions with obstacles and other agents. Specifically, if the $i$--th agent collides ($\lVert \boldsymbol{p}_i-\boldsymbol{p}_j\rVert < 2\rho_2$ or $\lVert \boldsymbol{p}_i-\boldsymbol{p}_{o,c}\rVert < \rho_2+\rho_3$), the avoidance reward $r_{i,a}=-1$. Otherwise, $r_{i,a} = 0$. This reward, therefore, actively promotes collision avoidance and safe navigation.
Although avoiding collisions is crucial, agent pathfinding must also prioritize minimizing path length to ensure efficiency. The upper layer introduces an intrinsic reward $r_{i,b}$, which is defined as follows:
\begin{align} 
\displaystyle r_{i,b} = -\frac{\lVert \boldsymbol{q}_i\left(\left\lfloor t / h \right\rfloor\right) - \boldsymbol{p}_i\left(t\right)\rVert}{\lVert \boldsymbol{q}_i\left(\left\lfloor t / h \right\rfloor\right) - \boldsymbol{p}_i\left(\left\lfloor t / h \right\rfloor\right) \rVert}. 
\label{equ8} 
\end{align}

To combine both the avoidance and path minimization objectives, we define the total reward in the lower layer as the sum of the avoidance and intrinsic rewards:
\begin{align} R_l\left( \boldsymbol{s}_{m+n},\boldsymbol{a}_n,\boldsymbol{q}_{m} \right) = \sum_i^N (\epsilon_1 r_{i,a}+ \left( 1 - \epsilon_1 \right)r_{i,b}). \label{equ9} \end{align}
where $\epsilon_1$ is the weighting factor that balances avoidance reward and intrinsic reward. It encourages the agent to navigate safely while minimizing the path length, facilitating efficient completion of its target--reaching task.

\subsection{Upper Layer for Task Allocation}
\label{sec:sample3B}
The goal $\boldsymbol{g}$ of the confrontation is to eliminate as many enemies as possible. Given the limited observations in the confrontation scenario, the local observation of the $i$--th agent includes information about observable allies and enemies, expressed as
\begin{align} \boldsymbol{z}_{u,i}=\left(\boldsymbol{u}_i,\boldsymbol{u}_{a,j},\boldsymbol{u}_{e,k}\right) ,
\label{equ10}
\end{align}
where $j,k \in \mathbb{N}_i$. Furthermore, the tasks of enemies $\boldsymbol{u}_{e,k}$ are unknown to the $i$--th agent.

\begin{figure}[t]
    \centering
    \includegraphics[width=0.4\textwidth]{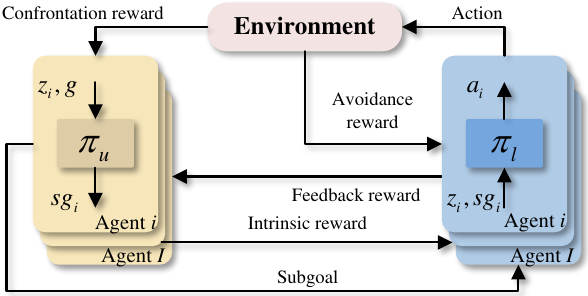}
    \caption{Interaction mechanism of two--layer networks.}
   \label{fig3}
\end{figure}

Based on the goal and observation, the agent chooses to either support an ally that is escaping or chasing, or to chase an enemy at a relative disadvantage. The upper layer generates the subgoal for the $i$--th agent by determining the target position defined as follows:
\begin{align} 
\boldsymbol{q}_{i}=\boldsymbol{p}_{a,j} ~ \text{or} ~ \boldsymbol{p}_{e,k} ,
\label{equ11}
\end{align}
where $f_{a,j}=-2$ or $f_{a,j}=0$, and $Adv\left(\boldsymbol{u}_i, \boldsymbol{u}_{e,k} \right) > 0$. The advantage function $Adv$ depends on the positions and velocities of $\boldsymbol{u}_i$ and $\boldsymbol{u}_{e,k}$, denoted as $
Adv\left(\boldsymbol{u}_i, \boldsymbol{u}_{e,k} \right) = \cos \theta_{i,k}$, where $\theta_{i,k}$ represents the angle between the vectors $\boldsymbol{v}_i$  and $\boldsymbol{p}_{e,k} - \boldsymbol{p}_i$. The selection of enemies and allies determines whether the agent is engaged in a chasing or supporting task. If neither allies nor enemies, as described above, are within the communication and perception ranges, the agent will perform a searching or escaping task and invoke the trajectory prediction model. After the lower layer executes $h$ steps, we feed the cumulative rewards to the upper layer, which adjusts its strategy based on the performance of the lower layer through bidirectional feedback. Therefore, the reward function of the upper layer in (\ref{equ1}) is improved as
\begin{equation}
    \begin{array}{c}
        \displaystyle R_u\left( \tau_m \right) = r_{\text{env}}\left( \boldsymbol{s}_{m+n},\boldsymbol{q}_m,\boldsymbol{g} \right) + \sum_{n=0}^{h-1}{ R_l\left( \boldsymbol{s}_{m+n},\boldsymbol{a}_n,\boldsymbol{q}_{m} \right)},
    \end{array}\label{equ13}
\end{equation}
where $r_{\text{env}}$ is the confrontation reward, defined as the total number of surviving allies and defeated enemies during these $h$ steps. The interaction mechanism between the lower and upper layers is illustrated in Fig. \ref{fig3}.

\subsection{Embedded Trajectory Prediction Model}
\label{sec:sample3C}
When the agent performs a searching or escaping task, it invokes a trajectory prediction model to compute the subgoal. In this paper, we assume that a global planner stores the historical trajectories of all enemies $p_{e,k}\left(t\right),t=0,\dots,T_n$, where $T_n$ represents the current time step. The planner then makes a preliminary trajectory prediction using a second--order dynamic model
\begin{equation}
\begin{array}{c}
\boldsymbol{p}_{e,k}\left(T_n+1\right)=\boldsymbol{p}_{e,k}\left(T_n\right)+\delta t \boldsymbol{v}_{e,k}\left(T_n\right)+\frac{1}{2}\delta t^2\boldsymbol{\dot{v}}_{e,k}\left(T_n\right),
\end{array}\label{equ14}
\end{equation}
where $\boldsymbol{\dot{v}}_{e,k}\left(T_n\right)=\left(\boldsymbol{v}_{e,k}\left(T_n\right)-\boldsymbol{v}_{e,k}\left(T_n-1\right) \right) / \delta t$.
To avoid collisions with obstacles in the confrontation scenario, the planner performs collision detection for each trajectory and further refines the trajectory through the potential field method. We consider an obstacle circle $c \in \mathbb{N}_k$ if it satisfies $\lVert \boldsymbol{p}_{e,k}\left(T_n+1\right) - \boldsymbol{p}_{o,c} \rVert < \rho_2+\rho_3$, where $\mathbb{N}_k$ denotes the set of obstacles that collide with the enemy $k$. The predicted position is then adjusted as follows:
\begin{align}
\boldsymbol{p}_{e,k}\left(T_n+1\right)&=\boldsymbol{p}_{e,k}\left(T_n+1\right) \nonumber \\ 
&+\delta t  \sum_{c\in \mathbb{N}_k}{\frac{ \boldsymbol{p}_{e,k}\left(T_n+1\right) - \boldsymbol{p}_{o,c}}{\lVert \boldsymbol{p}_{e,k}\left(T_n+1\right) - \boldsymbol{p}_{o,c} \rVert}d_3},
\label{equ15}
\end{align}
where $d_3=\rho_2+\rho_3-\lVert \boldsymbol{p}_{e,k}\left(T_n+1\right) - \boldsymbol{p}_{o,c} \rVert$. After iterating until time step $T_n + T_p$, the planner employs a hierarchical clustering algorithm \cite{pang2023cross} to group all predicted trajectory points. To generate trajectories for clustering, we repeatedly roll out the prediction model from each enemy’s most recent position. The resulting positions are then used for hierarchical clustering. The center of the $b_1$--th cluster, representing one of the most likely locations of enemies, is given by
\begin{align}
\displaystyle \boldsymbol{p}_{\text{cluster},b_1}=\frac{1}{N_p}\sum_{n=1}^{N_p}{\boldsymbol{p}_n},
\label{equ16}
\end{align}
where $\boldsymbol{p}_n$ is the position of each point in the cluster, and $N_p$ is the number of points in the cluster. The symbol $T_p$ is a constant that indicates the length of the prediction time step. If there are no enemies within the perception range, and no allies in chasing or escaping within the communication range, the $i$--th agent will switch to a searching task, selecting the nearest center as its subgoal
\begin{align}
\displaystyle \boldsymbol{q}_{i}=\underset{\boldsymbol{p}_{\text{cluster},b_1}}{\arg\min} ~ \lVert \boldsymbol{p}_{\text{cluster},b_1} - \boldsymbol{p}_{i} \rVert.
\label{equ17}
\end{align}
Otherwise, if an enemy is within the perception range of the $i$--th agent and satisfies $Adv\left(\boldsymbol{u}_i, \boldsymbol{u}_{e,k} \right) < 0$, the agent will perform an escaping task. To compute the least likely locations of enemies, the planner randomly generates several candidate points that satisfy the following conditions:
\begin{align}
\displaystyle \lVert \boldsymbol{p}_{\text{candidate},b_2}-\boldsymbol{p}_{\text{cluster},b_1} \rVert > d_4, \forall b_1,\forall b_2,
\label{equ18}
\end{align}
where $d_4$ is a constant indicating the distance of the candidate points from the most likely locations. The subgoal of the escaping agent can be expressed as
\begin{align}
\displaystyle \boldsymbol{q}_{i}=\underset{\boldsymbol{p}_{\text{candidate},b_2}}{\arg\min} ~ \lVert \boldsymbol{p}_{\text{candidate},b_2} - \boldsymbol{p}_{i} \rVert.
\label{equ19}
\end{align}
The subgoal is then transmitted from the global planner to the $i$--th agent.

\subsection{Training Methods for Two Layers}
\label{sec:sample3D}
Combining the goal--conditioned MDP framework, we adopt double DQN for training in the upper layer. In this approach, each agent estimates the optimal state--action value function $Q_i^*:\mathbb{S}\times \mathbb{A}\times \mathbb{G}\rightarrow \mathbb{R}$ through a parameterized neural network $Q_{\phi_{u,i}}\left( \boldsymbol{s}_i\left(t\right),\boldsymbol{a}_i\left(t\right), \boldsymbol{g} \right)$. The subscript $\phi_u=\left(\phi_{u,1},\dots,\phi_{u,N}\right)$ represents the weight parameters of the value--function network in the upper layer. For $\gamma \approx 1$, $Q^*$ estimates the discounted returns of the optimal strategy over an infinite horizon. The method approximates $Q^*$ using the network $Q_{\phi_{u,i}}$, and the loss function is formulated as
\begin{equation}
    \begin{array}{c}
        L\left( \phi_{u,i} \right)=E_{ \boldsymbol{s},\boldsymbol{a},\boldsymbol{g} }\lVert Q_{\phi_{u,i}}\left( \boldsymbol{s}_i\left(t\right),\boldsymbol{a}_i\left(t\right), \boldsymbol{g} \right) -y_i \rVert ^2,
    \end{array}\label{equ20}
\end{equation}
where $E$ denotes the mathematical expectation. The $Q$--target $y_i$ is defined as
\begin{align}
y_i&=R_u\left( \boldsymbol{s}_i\left(t\right),\boldsymbol{a}_i\left(t\right), \boldsymbol{g} \right) \nonumber \\
&+\gamma \max Q_{\phi_{u,i}}\left( \boldsymbol{s}_i\left(t+h\right),\boldsymbol{a}_i\left(t+h\right), \boldsymbol{g} \right).
\label{equ21}
\end{align}

To enable agents to make decisions based on local observations while collaborating with each other to achieve subgoals in the lower layer, we use MADDPG, a method that employs centralized training with decentralized execution. It designs a separate critic network $Q_{\phi_{l,i}}\left( \boldsymbol{s}\left(t\right),\boldsymbol{a}\left(t\right), \boldsymbol{q}\left(\lfloor t / h \rfloor\right) \right)$ for each agent, which is updated as follows:
\begin{equation}
    \begin{array}{c}
L\left( \phi_{l,i} \right)=E_{\boldsymbol{s},\boldsymbol{a},\boldsymbol{q} }\lVert Q_{\phi_{l,i}}\left( \boldsymbol{s}\left(t\right),\boldsymbol{a}\left(t\right), \boldsymbol{q}\left(\left\lfloor t / h \right\rfloor\right) \right) -y_{i} \rVert ^2,
    \end{array}\label{equ23}
\end{equation}
where the subscript $\phi_l=\left(\phi_{l,1},\dots,\phi_{l,N}\right)$ represents the weight parameters of the critic network in the lower layer. The $Q$--target is revised as
\begin{align}
  y_i=&R_l\left( \boldsymbol{s}_i\left(t\right),\boldsymbol{a}_i\left(t\right), \boldsymbol{q}_i\left(\left\lfloor \displaystyle t / h \right\rfloor\right) \right) \nonumber \\
&+\gamma \max Q_{\phi_{l,i}}\left( \boldsymbol{s}\left(t+1\right),\boldsymbol{a}\left(t+1\right), \boldsymbol{q}\left(\left\lfloor  \displaystyle  t / h \right\rfloor + 1\right) \right).
\label{equ24}
\end{align}
In addition, each agent maintains a policy network $\pi_{\omega_{l,i}}\left(\boldsymbol{z}_i\left(t\right),\boldsymbol{q}_i\left(\lfloor t / h \rfloor\right)\right)$ to determine an action based on its local observation, with the following loss function:
\begin{equation}
    \begin{array}{c}
        L\left( \omega_{l,i} \right)=E_{\boldsymbol{s},\boldsymbol{a},\boldsymbol{q} } \left(- Q_{\phi_{l,i}}\left( \boldsymbol{s}\left(t\right),\boldsymbol{a}\left(t\right), \boldsymbol{q}\left( \left\lfloor t / h \right\rfloor \right) \right)\right).
    \end{array}\label{equ25}
\end{equation}
The subscript $\omega_l=\left(\omega_{l,1},\dots,\omega_{l,N}\right)$ represents the weight parameters of the policy network in the lower layer. 

To ensure stable policy convergence under the strong coupling between the two layers, we introduce a cross--training method. The training alternates between updating the lower--layer policies every \textit{h} steps and subsequently refining the upper--layer policy using the accumulated feedback. Details are outlined in Algorithm \ref{alg1}. This method updates the upper and lower networks simultaneously, enabling both layers to adjust their policies in real time.

\begin{algorithm}[t]
    \renewcommand{\algorithmicrequire}{\textbf{Input:}}
    \renewcommand{\algorithmicensure}{\textbf{Initialization:}}
	\caption{Cross--Training Method}
    \label{alg1}
    \begin{algorithmic}[1] 
        \REQUIRE  the number of training steps $T$; the number of cross--training episodes $E_c$;  the length of  each subtask $h$;
	    \ENSURE the parameters of the networks in the upper layer $\phi_u$ and lower layer $\phi_l,\omega_l$; 
        
        \FORALL {$e\ =\ 1\rightarrow E_c$}
            \STATE $t_0 \gets 0$
            \FORALL {$t\ =\ 0\rightarrow T$}
                \STATE $R_l \gets$ Eq. (\ref{equ9});
                \STATE $L\left(\phi_l\right) \gets$ Eq. (\ref{equ23}), $L\left(\omega_l\right) \gets$ Eq. (\ref{equ25});
                \STATE Refresh networks $Q_{\phi_l}, \pi_{\omega_l}$;
                \COMMENT{Lower layer training}
                \IF{$t = t_0 + h$}
                    \STATE $R_u \gets$ Eq. (\ref{equ13});
                    \STATE $L\left(\phi_u\right) \gets$ Eq. (\ref{equ20});
                    \STATE Refresh network $Q_{\phi_u}$;
                    \COMMENT{Upper layer training}
                    \STATE $t_0 \gets t$
                \ENDIF
            \ENDFOR
        \ENDFOR
    \end{algorithmic}
\end{algorithm}

\section{Experiments}
\label{sec:sample4}
In this section, extensive experiments are conducted to evaluate the performance of our HRL method in solving the strategic confrontation problem at different sizes. In comparative experiments, we adopt various baselines, including the expert system, heuristic algorithm, and single--layer DRL. We assess the impact of the interaction mechanism through ablation studies. Moreover, we test the model trained on a small scale to solve larger--scale problems, investigating the generalization of our method. Finally, we deploy the cross--trained model from simulations to the real--robot system.
\subsection{Setting Up}
\label{sec:sample4A}
Before analyzing the comparative results, we outline the detailed settings used in the simulations. Following the conventions in \cite{hou2023hierarchical,nian2024large}, we consider an equal number of red and blue agents, randomly distributed in a 30 m$\times$20 m scenario with static obstacles. In our experiment, three scenarios are considered: five agents versus five agents (denoted V5), seven agents versus seven agents (denoted V7), and nine agents versus nine agents (denoted V9). The attack radius $\rho_1$, angle $\theta_1$, and probability $\epsilon$ are set to 1 m, $\pi /$2, and 0.5, respectively. The avoidance radius $\rho_2$ and the obstacle radius $\rho_3$ are set to 0.3 m. The observation length $d_1$ and width $d_2$ are set to 5 m. The maximum velocity magnitude $\lVert \boldsymbol{v} \rVert_{\text{max}}$ is 2 m/s, and the step length $\delta t$ is 0.1 s. 

For algorithm training, we cross--train the two layers for $E_c =$ 60 episodes and $T=$ 300 steps in 100 randomly generated instances. The networks in both layers are implemented as three--layer MLPs with hidden units [128, 64, 32]. All networks use ReLU activations and Adam optimizer. The replay buffer size is 10$^5$, and mini--batch size is 64. The discount factor $\gamma$ and the learning rate $\alpha$ are set to $0.99$ and $0.001$, respectively. To plot the experimental curves, we use solid curves to represent the mean values across all instances, with shaded regions indicating the standard deviation. In our experiments, we apply the HRL method and baselines to the blue agents, while the red agents use traditional TAMP methods, including the expert system \cite{hou2023hierarchical} and heuristic algorithm \cite{liu2022evolutionary}.

\subsection{Comparative Analysis}
\label{sec:sample4B}
In cross--training, we employ the HRL method to enhance the performance of both layers. We compare this method with baselines, including the expert system, heuristic approach, and single--layer DRL. Additionally, to investigate the influence of interactions between the two layers, we conduct an ablation study by removing the feedback reward in HRL. The baselines are briefly introduced as follows.

\begin{itemize}
\item[1)] Expert system: Based on the confrontation rules constructed in \cite{hou2023hierarchical}, this algorithm uses a hierarchical decision--making framework to match optimal task allocation and path planning. 
\item[3)] Heuristic algorithm: The algorithm in \cite{liu2022evolutionary} designs an evolutionary--based attack strategy for swarms, evaluating the potential benefits or threats of the next perceptible attack positions of an agent. It also incorporates a collision--avoidance method for feasible path planning.
\item[4)] Single--layer DRL: The algorithm in \cite{qu2023pursuit} treats confrontation as an MDP process and employs single--layer DRL to output the agents' strategies end--to--end.
\item[5)] HRL/FB: This approach adopts our HRL method but updates the upper layer without the feedback reward transferred from the lower layer.
\end{itemize}

\begin{figure*}[t]
    \centering
    \includegraphics[width=0.9\textwidth]{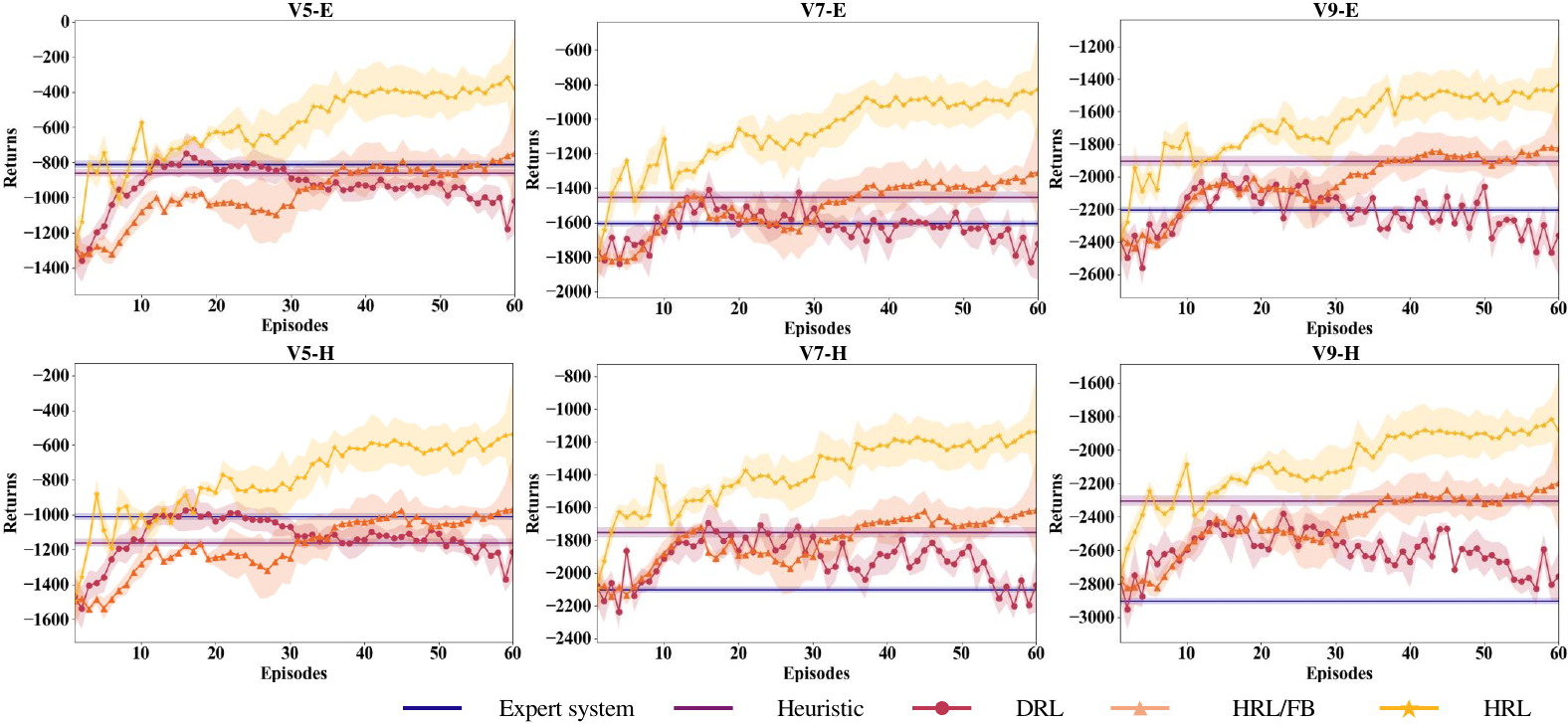}
    \caption{Learning curves of baselines and our method for different swarm sizes.}
    \label{fig4}
\end{figure*}

\begin{table*}[t]
\renewcommand{\arraystretch}{1.06}
\setlength{\tabcolsep}{7pt}
\centering
\caption{Experiment Results of Baselines and Our Method for Different Swarm Sizes.}
\label{table1}
\begin{tabular}{|c|c|c|c|c|c|c|c|c|c|c|c|c|}
\hline
\multicolumn{2}{c|}{Algorithm of} & \multicolumn{2}{c|}{\multirow{2}{*}{Method}} &\multicolumn{3}{c|}{V5} &\multicolumn{3}{c|}{V7} & \multicolumn{3}{c}{V9}\\

\multicolumn{2}{c|}{enemies} & \multicolumn{2}{c|}{} & \multicolumn{1}{c}{Re.} & \multicolumn{1}{c}{Ti.($\times 10^{-2}$ s)} & \multicolumn{1}{c|}{W.R.(\%)} & \multicolumn{1}{c}{Re.} & \multicolumn{1}{c}{Ti.($\times 10^{-2}$ s)} & \multicolumn{1}{c|}{W.R.(\%)} & \multicolumn{1}{c}{Re.} & \multicolumn{1}{c}{Ti.($\times 10^{-2}$ s)} & \multicolumn{1}{c}{W.R.(\%)}\\
\hline

\multicolumn{2}{c|}{\multirow{5}{*}{Expert}} & \multicolumn{2}{c|}{Expert} & \multicolumn{1}{c}{-1009}	&\multicolumn{1}{c}{3.21}	&\multicolumn{1}{c|}{51}	&\multicolumn{1}{c}{-1602}	&\multicolumn{1}{c}{3.41}	&\multicolumn{1}{c|}{50}	&\multicolumn{1}{c}{-2202}	&\multicolumn{1}{c}{3.70}	&\multicolumn{1}{c}{51}\\ 

\multicolumn{2}{c|}{} & \multicolumn{2}{c|}{Heuristic} &\multicolumn{1}{c}{-1159}	&\multicolumn{1}{c}{2.67}	&\multicolumn{1}{c|}{48}	&\multicolumn{1}{c}{-1452}	&\multicolumn{1}{c}{4.26}	&\multicolumn{1}{c|}{52}	&\multicolumn{1}{c}{-1902}	&\multicolumn{1}{c}{6.07}	&\multicolumn{1}{c}{54}\\ 

\multicolumn{2}{c|}{} & \multicolumn{2}{c|}{DRL} &\multicolumn{1}{c}{-1214}	&\multicolumn{1}{c}{\textbf{0.26}}	&\multicolumn{1}{c|}{46}	&\multicolumn{1}{c}{-1723}	&\multicolumn{1}{c}{\textbf{0.32}}	&\multicolumn{1}{c|}{47}	&\multicolumn{1}{c}{-2362}	&\multicolumn{1}{c}{\textbf{0.45}}	&\multicolumn{1}{c}{50}\\

\multicolumn{2}{c|}{} & \multicolumn{2}{c|}{HRL/FB} &\multicolumn{1}{c}{-967}	&\multicolumn{1}{c}{\multirow{2}{*}{0.53}}	&\multicolumn{1}{c|}{53}	&\multicolumn{1}{c}{-1309}	&\multicolumn{1}{c}{\multirow{2}{*}{0.69}}	&\multicolumn{1}{c|}{54}	&\multicolumn{1}{c}{-1825}	&\multicolumn{1}{c}{\multirow{2}{*}{0.95}}	&\multicolumn{1}{c}{56}\\

\multicolumn{2}{c|}{} & \multicolumn{2}{c|}{HRL} &\multicolumn{1}{c}{\textbf{-538}}	&\multicolumn{1}{c}{}	&\multicolumn{1}{c|}{\textbf{82}}	&\multicolumn{1}{c}{\textbf{-827}}	&\multicolumn{1}{c}{}	&\multicolumn{1}{c|}{\textbf{84}}	&\multicolumn{1}{c}{\textbf{-1433}}	&\multicolumn{1}{c}{}	&\multicolumn{1}{c}{\textbf{85}}\\
\hline

\multicolumn{2}{c|}{\multirow{5}{*}{Heuristic}} & \multicolumn{2}{c|}{Expert} & \multicolumn{1}{c}{-809}	&\multicolumn{1}{c}{3.22}	&\multicolumn{1}{c|}{52}	&\multicolumn{1}{c}{-2102}	&\multicolumn{1}{c}{3.40}	&\multicolumn{1}{c|}{48}	&\multicolumn{1}{c}{-2902}	&\multicolumn{1}{c}{3.73}	&\multicolumn{1}{c}{46}\\ 

\multicolumn{2}{c|}{} & \multicolumn{2}{c|}{Heuristic
} &\multicolumn{1}{c}{-859}	&\multicolumn{1}{c}{2.66}	&\multicolumn{1}{c|}{49}	&\multicolumn{1}{c}{-1752}	&\multicolumn{1}{c}{4.28}	&\multicolumn{1}{c|}{51}	&\multicolumn{1}{c}{-2302}	&\multicolumn{1}{c}{6.06}	&\multicolumn{1}{c}{50}\\ 

\multicolumn{2}{c|}{} & \multicolumn{2}{c|}{DRL
} &\multicolumn{1}{c}{-1018}	&\multicolumn{1}{c}{\textbf{0.27}}	&\multicolumn{1}{c|}{46}	&\multicolumn{1}{c}{-2072}	&\multicolumn{1}{c}{\textbf{0.30}}	&\multicolumn{1}{c|}{49}	&\multicolumn{1}{c}{-2754}	&\multicolumn{1}{c}{\textbf{0.47}}	&\multicolumn{1}{c}{47}\\

\multicolumn{2}{c|}{} & \multicolumn{2}{c|}{HRL/FB
} &\multicolumn{1}{c}{-747}	&\multicolumn{1}{c}{\multirow{2}{*}{0.52}}	&\multicolumn{1}{c|}{59}	&\multicolumn{1}{c}{-1617}	&\multicolumn{1}{c}{\multirow{2}{*}{0.71}}	&\multicolumn{1}{c|}{54}	&\multicolumn{1}{c}{-2196}	&\multicolumn{1}{c}{\multirow{2}{*}{0.99}}	&\multicolumn{1}{c}{52}\\

\multicolumn{2}{c|}{} & \multicolumn{2}{c|}{HRL} &\multicolumn{1}{c}{\textbf{-381}}	&\multicolumn{1}{c}{}	&\multicolumn{1}{c|}{\textbf{84}}	&\multicolumn{1}{c}{\textbf{-1138}}	&\multicolumn{1}{c}{}	&\multicolumn{1}{c|}{\textbf{82}}	&\multicolumn{1}{c}{\textbf{-1881}}	&\multicolumn{1}{c}{}	&\multicolumn{1}{c}{\textbf{81}}\\
\hline
\end{tabular}
\end{table*}

To ensure a fair comparison between our method and the baselines, we evaluate all methods under the same experimental conditions, using identical input data and performance metrics. The HRL learning curves and the baselines for different swarm sizes are shown in Fig. \ref{fig4}, where "VX--E" and "VX--H" represent the use of the expert system and heuristic algorithm as enemies' strategies in the VX scenario, respectively. Both the expert system and heuristic algorithm can develop effective strategies for agents across different instances. Between these two conventional TAMP algorithms, the heuristic algorithm outperforms the expert system in large--scale confrontations.

Initially, learning--based algorithms underperform due to the randomness of their starting strategies. With continuous exploration and training, their performance gradually surpasses that of the non--learning algorithms. Due to the hybrid decision--making process in the confrontation, the performance of single--layer DRL deteriorates over time during training. In contrast, HRL consistently finds better strategies than non--learning algorithms across various swarm sizes. Furthermore, when the feedback reward is omitted, the performance curve converges to a suboptimal value due to the lack of effective interaction between the two layers.

In addition, we deploy the policy networks trained using learning--based methods across another 100 instances and compare their performance with conventional TAMP algorithms. To provide a more comprehensive assessment, we introduce two additional evaluation metrics: decision time and confrontation win rate. Decision time refers to the average time that all agents take to allocate tasks and plan the path during each time step. Confrontation win rate measures the ratio of successful outcomes to the total number of instances. The average experimental results across 100 instances for different swarm sizes are summarized in Table \ref{table1}, which presents the episode returns (Re.), decision time (Ti.), and confrontation win rate (W.R.) for all methods.

Notably, all methods support real--time decision--making at each step, with decision time below the time interval $\delta t$.
We observe a clear relationship between the win rate and episode returns, which are calculated using the reward function in this study. Agents train their strategies to maximize returns, which in turn leads to a higher win rate. 

Between the two non--learning algorithms, the heuristic algorithm has a longer decision time, though its returns and win rate improve as the problem size increases. Due to its end--to--end nature, DRL exhibits the shortest decision time, but its win rate is significantly lower than that of HRL. The results demonstrate that our method outperforms the baselines in terms of episode returns, decision time, and confrontation win rate, especially in large--scale instances. This is attributed to the effectiveness of our approach in managing interactions between the upper and lower layers by decoupling confrontation into allocation and planning.

\begin{figure}[t]
    \centering
    \includegraphics[height=0.547\textwidth]{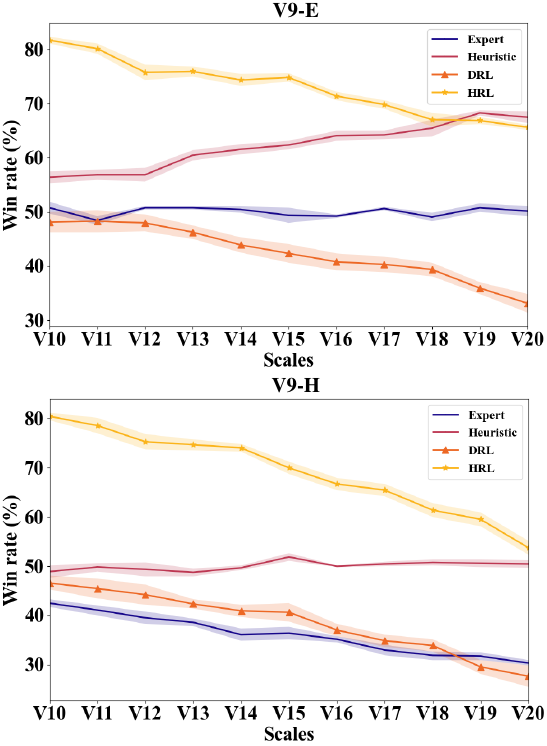}
\caption{Generalization to larger--scale instances.}
\label{fig5}
\end{figure}

\begin{figure}[t]
    \centering
    \includegraphics[height=0.547\textwidth]{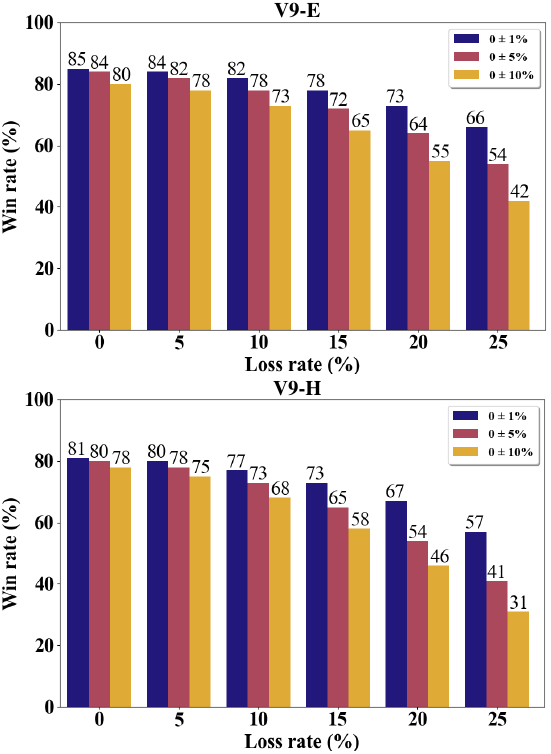}
\caption{Assessment in instances involving communication and sensor uncertainties.}
\label{fig5_0}
\end{figure}

\subsection{Generalization to Larger--Size Instances}
\label{sec:sample4C}
In large--scale confrontation scenarios, retraining strategies can lead to significant consumption of computational resources. Therefore, our goal is for the trained models to exhibit strong generalization performance. In our method, the decentralized policies of agents can be applied directly to additional agents without the need for retraining. To assess the generalization capability of our approach, we apply a set of neural network parameters, trained on the V9 scenario, to instances involving a higher number of agents (ranging from 10 to 20). As the scenario size increases, agents face greater challenges and experience lower win rates. However, our method continues to maintain favorable solution quality compared to the baselines.

The generalization results are shown in Fig. \ref{fig5}, which presents the win rate across different instance scales. Due to the limited generalization capabilities of traditional TAMP algorithms, they require repeated adjustments for larger--scale scenarios. As shown, our method outperforms expert system, heuristic algorithm, and single--layer DRL in most instances. Our approach demonstrates high scalability in larger scenarios compared to the baselines. In particular, the model trained on the V9 scenario can generalize to V20, achieving win rates consistent with those achieved by the heuristic algorithm reiterated on V20. These results verify that policies trained on small--scale instances can effectively make decisions in scenarios of varying scales. Thus, our method exhibits favorable generalization capability for the strategic confrontation problem.

\subsection{Deployment in Real--Robot System}
\label{sec:sample4D}
Before presenting the real--robot experiments, we evaluate the performance of our method through simulations that account for communication and sensor uncertainties, including message loss between neighbors and sensor noise error. The results are shown in Fig. \ref{fig5_0}, where the horizontal axis represents the message loss rate, the vertical axis indicates the win rate, and the legend represents different levels of Gaussian noise error. Although the success rate decreases as the uncertainty increases, it remains above 73\% when the loss rate is below 10\% and the noise error is within ±5\%. These results demonstrate that our method is suitable for real--world applications. 

\begin{figure*}[t]
    \centering
    \includegraphics[width=0.9\textwidth]{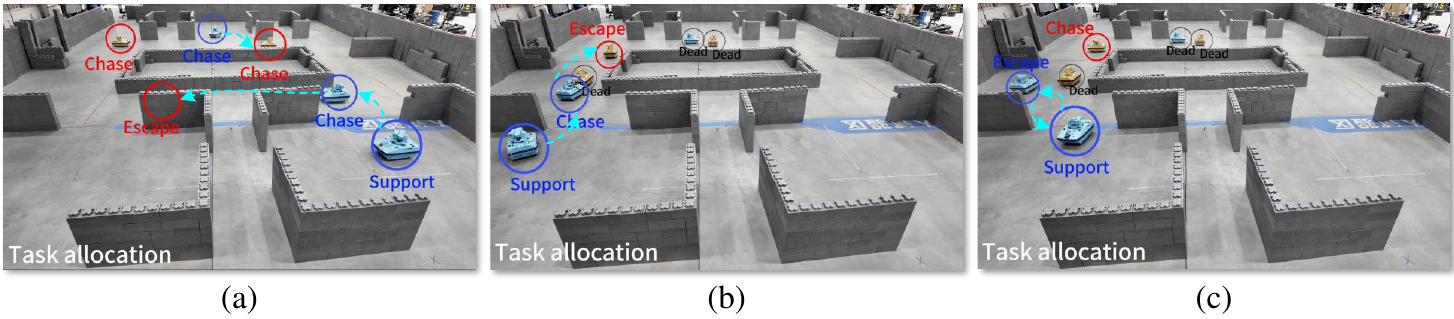}
    \caption{The decision--making process for V3 scenario in the real--robot system.}
    \label{fig6}
\end{figure*}

We then conduct experiments using an actual robotic system, as depicted in Fig. \ref{fig6}. A V3 confrontation is considered in a static--obstacle scenario, with experimental settings consistent with those defined in Section \ref{sec:sample2A}. By performing cross--training in simulations, we obtain two--layer networks for the agents in this scenario. During the experiment, we control the agents via a ground control system, with motion capture provided by Optitrack. Our algorithm receives data from Optitrack that includes the positions and velocities of all agents. The algorithm subsequently calculates observations from perception sensors and generates position commands based on the two--layer networks, which are sent to the ground control system. Using these networks, the agents successfully defeat all enemies during the experiments while avoiding collisions. After sufficient training with our method, the agents exhibit coordinated confrontation behaviors, such as prioritizing support for teammates in critical situations (Fig. \ref{fig6}(a)), using static obstacles for cover (Fig. \ref{fig6}(b)), and luring opponents into teammates' encirclement (Fig. \ref{fig6}(c)).

\section{Conclusions}
\label{sec:sample5}
This paper has presented a bidirectional HRL approach to address the hybrid decision--making process in strategic confrontations. It has developed two--layer DRL networks that associate commands and actions with task allocation and path planning, respectively. Furthermore, the method has refined the interaction mechanism and introduced a cross--training method to enhance learning between the two layers. We also have incorporated a trajectory prediction model to convert abstract command representations into concrete action objectives. Experimental results have demonstrated that our method outperforms the baselines in terms of episode returns, decision time, and confrontation win rate. Finally, the generalization and adaptability of our method have been validated by applying the trained model to larger instances and real--robot systems, respectively.
\bibliographystyle{IEEEtran}
\bibliography{iros}

\end{document}